\RequirePackage{rotating}

\documentclass[sigconf]{acmart}

\AtBeginDocument{%
  \providecommand\BibTeX{{%
    \normalfont B\kern-0.5em{\scshape i\kern-0.25em b}\kern-0.8em\TeX}}}

\setcopyright{none}
\usepackage[pdf]{graphviz}
\usepackage{graphicx}
\usepackage{amsmath}
\usepackage{array}
\usepackage{wrapfig}
\usepackage{color}
\usepackage{rotating}
\usepackage{tabularx}
\usepackage[section]{placeins}

\newcolumntype{L}{>{\arraybackslash}m{4cm}}
\DeclareMathOperator*{\argmax}{arg\,max}

\acmConference[KDD2022]{28TH ACM SIGKDD CONFERENCE ON KNOWLEDGE DISCOVERY AND DATA MINING}{August 14-18, 2022}{Washington DC}

\acmPrice{15.00}
\acmISBN{978-1-4503-XXXX-X/18/06}

\begin{document}

\title{ML Prescriptive Canvas for Optimizing Business Outcomes}

\author{Hanan Shteingart}
\authornote{Both authors contributed equally to this research.}
\affiliation{%
    \institution{VIANAI Systems}
    \country{Israel}
    \city{Herzliya}
    }
\email{hanan@vian.ai}

\author{Gerben Oostra}
\authornotemark[1]
\affiliation{%
    \institution{VIANAI Systems}
    \country{Israel}
    \city{Herzliya}
    }
\email{gerben@vian.ai}

\author{Ohad Levinkron}
\affiliation{
 \institution{VIANAI Systems}
  \country{Israel}
  \city{Herzliya}
  }
\email{ohad@vian.ai}

\author{Naama Parush}
\affiliation{
 \institution{VIANAI Systems}
  \country{Israel}
  \city{Herzliya}
  }
\email{naama@vian.ai}

\author{Gil Shabat}
\affiliation{%
  \institution{VIANAI Systems}
  \country{Israel}
  \city{Herzliya}
  }
\email{gil@vian.ai}

\author{Daniel Aronovich}
\affiliation{%
 \institution{VIANAI Systems}
   \country{Israel}
   \city{Herzliya}
   }
\email{aronovich.daniel@vian.ai}

\renewcommand{\shortauthors}{Shteingart and Oostra, et al.}



\begin{abstract}
Data science has the potential to improve business in a variety of verticals. While the lion's share of data science projects uses a predictive approach, to drive improvements these predictions should become decisions. However, such a two-step approach is not only sub-optimal but might even degrade performance and fail the project. The alternative is to follow a prescriptive framing, where actions are ``first citizens'' so that the model produces a policy that prescribes an action to take, rather than predicting an outcome. In this paper, we explain why the prescriptive approach is important and provide a step-by-step methodology: the Prescriptive Canvas. The latter aims to improve framing and communication across the project stakeholders including project and data science managers towards a successful business impact.

\end{abstract}




\maketitle



\section{Predictive vs Prescriptive}\label{sec:prescriptions}
Data science (DS) projects commonly aim to improve business decisions. In this section, we will explain the three-level analytics model that distinguishes between different tasks within data science. We will then show that the predictive approach is more common than the prescriptive one, and explain why the prescriptive approach is better when one aims to improve business decisions.
\subsection{The Three Levels of Analytics}
\begin{figure}[!htb]
    \centering
    \includegraphics[width=0.35\textwidth]{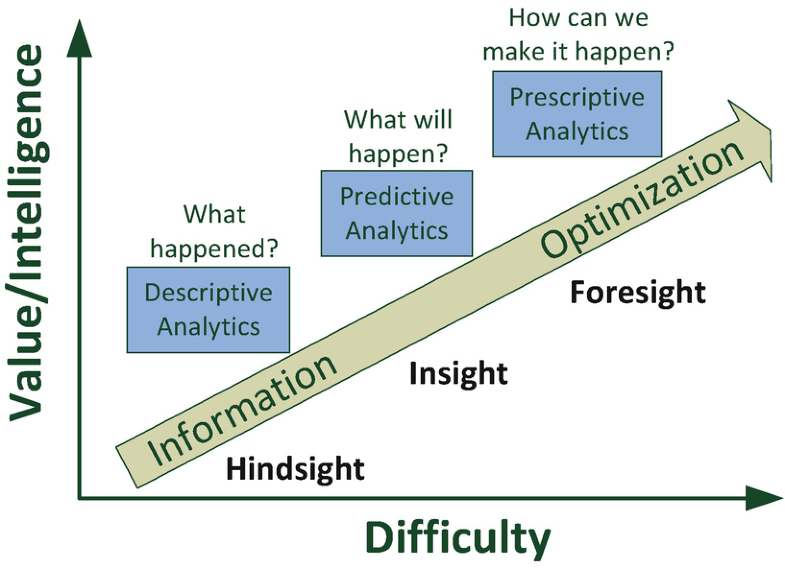}
    \Description{Shows the three stages, each as a block, in a chart comparing value/intelligence against difficulty. The first block, located at low difficulty and low value, is "Descriptive Analytics", annotated with "What happened?" and "Hindsight". The second block, located at medium difficulty and medium value, is "Predictive Analytics", annotated with "What will happen?" and "Insight". The third block, located at high difficulty and high value, is Prescriptive Analytics", annotated with "How can we make it happen?" and "Foresight". The upward diagonal is highlighted with an arrow, from Information at the tail (near Descriptive Analytics) to Optimization at the head (near Prescriptive Analytics).}
    \caption{Descriptive, predictive, and prescriptive analytics are three stages of business analytics, characterized by different levels of difficulty and value \cite{Siksnys2018}.}
    \label{fig:ladder}
\end{figure}

There are three levels of data science tasks \cite{hernan2019second, griffin2020big, Lo2020Top}, ordered by increasing value and difficulty, as depicted in Figure \ref{fig:ladder}:
\begin{itemize}
    \item \textbf{Descriptive}: \textit{What happened?}
        e.g.\ What was the average churn over the last 3 months? The result is often a value, chart or dashboard.
    \item \textbf{Predictive}: \textit{What will happen?}
        e.g.\ What is the probability this user will churn? The result is a \textit{model} that is able to provides predictions.
    \item \textbf{Prescriptive}: \textit{What to do?}
        e.g.\ What should we do to prevent churn? The result is a model that represents a \textit{policy}: it prescribes actions that optimize the outcome.
\end{itemize}

The first and the second questions are well understood and even can be automated in some cases\cite{he2021automl}. However, the last prescriptive question aims to directly optimize business value \cite{athey2017beyond}. Even though most data science projects aim to improve business decisions, they often solve a forecasting or prediction problem.

To demonstrate this, we analyzed the Kaggle meta-database\cite{kagglemeta} which describes 5,388 data science competitions. While "forecast" appears in all competition's descriptions, neither "action", "policy" nor "decision" appear even once. This again demonstrates the dominance of the predictive framing approach. As an exception, recommendation competitions appeared 11 times. Though the recommendation problem is prescriptive in nature, it is often wrongly framed as a predictive problem \cite{wang2020causal} and therefore tries to predict future choices or ratings rather than the effect of these recommendations, also known as uplift \cite{fang2018uplift}.


\subsection{Example: Predictive vs Prescriptive Churn}
To demonstrate the difference between predictive and prescriptive approaches, consider the classical problem of churn prevention. A predictive model provides expected churn probabilities, which can be used in a heuristic policy to drive actions, for example targeting users with the highest churn probability. Formally the policy would be $\pi^{\text{pred}}(x) = I(\hat{y}(x),t)$ where $t$ is some threshold, $y$ is churn probability and $I(a,b)=1$ if $a>b$, otherwise zero. However, the real business goal is to reduce churn, rather than predict it \cite{devriendt2021}. While the predictive approach targets users that are likely to churn, the prescriptive approach targets users that can be persuaded. Research has shown that the prescriptive approach could be up to $3x$ more effective than the predictive one\cite{devriendt2021, ascarza2018retention}. 

To explain why the prescriptive approach is better, consider that customers that are likely to churn aren't necessarily easily retained. Some users will be ``lost causes''; they will always churn despite any action, thus targeting these will be a waste of effort. Some other users will be ``sleeping dogs''; they would churn because of the action. Perhaps they were already thinking of canceling their subscription, but by contacting them, they churn at that moment. The prescriptive approach aims to correctly predict the effect of the action, and as such, prescribe actions that truly improve churn. Formally, the prescriptive policy would be $\pi^{\text{prescriptive}} = I(\tau(x), 0)$ where $\tau(x)$ is the estimated effect of the targeting (the difference in retention probability if targeted or not). We will discuss this kind of approach, called Uplift Modeling in section \ref{sec: uplift}.

Note that besides the different views on model output, this also implies choosing correctly the business objective. If a retention desk is measuring success by the amount of retention in the targeted population, a predictive approach would be optimal. However, the latter objective will encourage targeting the ``sure things'' rather than the ``persuadables'', thus wasting the targeting budget. However, if the correct business objective will be used, minimizing the overall churn rate, a prescriptive approach will be optimal and likely to impact the true business goals. 

\subsection{The Prescriptive Problem}
The predictive approach goal is to predict as accurate as possible. Formally, minimizing some loss function $\mathcal{L}(\hat{y},y)$ where $y$ is the true label and $\hat{y}(x)$ is the predicted output based on the input $x$. By contrast, the goal of prescriptive modelling is to find a policy that optimizes some reward by dictating actions. Formally, at each context $x$, the policy $Pr(a) = \pi(x)$ dictates which action $a$ should be taken at which probability, such that reward is maximized: $\pi^{\text{opt}} = \argmax_{\pi}\mathcal{R}(\pi)$.

The fundamental problem in prescriptive modeling is that we do not know what would have happened if we had acted differently, also known as the \textbf{counterfactual} outcome. In some sense, this could be interpreted as a missing data problem. If we could simulate using a ``what-if'' \cite{hernan2020causal} model what would have been the counterfactual, we could have used it to find the optimal policy. This also implies that measuring the performance of a prescriptive model offline is harder compared to the predictive approach. Because we don't know the true counterfactual outcomes, we can't use them to evaluate candidate policies, as we need to essentially solve the "what-if" problem. As a consequence also model-selection and hyper-parameter optimization are challenging. Nevertheless, there are several approaches to estimate the performance, also known as offline policy evaluation (OPE) \cite{levine2020offline, dudik2011doubly}, and there are some methods for model selection, e.g. \cite{schuler2017synth}. By contrast, in predictive modeling, model selection and hyper-parameter optimization are relatively straightforward via a holdout sample.

\begin{figure}[!ht]
    \centering
    
    
    \includegraphics[width=0.35\textwidth]{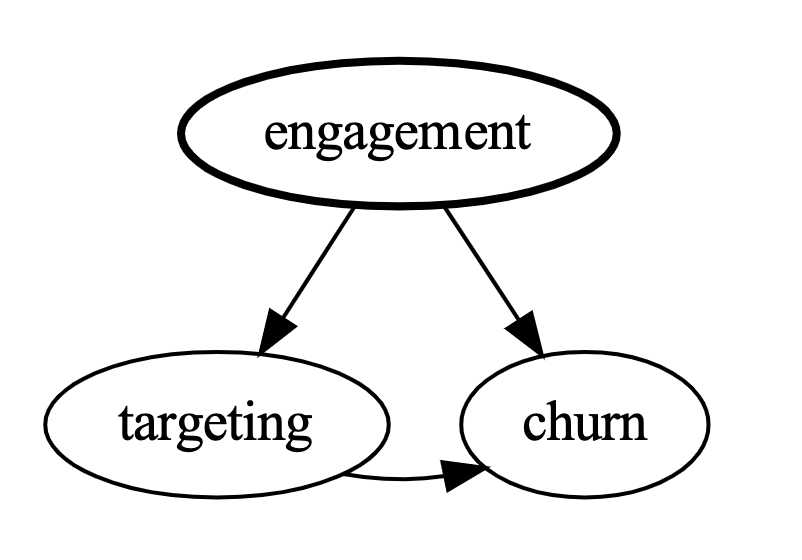}
    \caption{An example of a DAG in a marketing scenario. The engagement level affects both the targeting (treatment) and the churn probability (outcome), thus it is a confounder. If engagement is not controlled for, a biased effect of targeting on churn would be estimated: e.g. targeting causes churn rather than decreasing it. }
    \Description{A directed acyclic graph (DAG) with three nodes: engagement, targeting, and churn. The engagement has outgoing arrows to both targeting and churn. Targeting has an arrow to churn.}
    \label{fig:dag}
\end{figure}

To model the counterfactual we need to understand the causal relationship which can be learned from domain experts or using causal discovery techniques \cite{https://doi.org/10.48550/arxiv.2103.02582}.  Unfortunately, the data does not explicitly provide us the causal relationships but merely the correlations. How can we distinguish causation from correlation? Understanding the data generating process can help distinguish causation from correlation. This typically is represented in a directed acyclic graph (DAG) of causal dependencies. A simplified example is depicted in Figure \ref{fig:dag}. In this example, customers are contacted based on their engagement. However, their engagement level is also a precursor of their churn. The relationship between targeting and engagement to churn can be found in the data. In do-calculus language  ~\cite{pearl2018book}, there are two paths from targeting to churn. One is the direct causal dependency we are interested in, and another so-called back door path through engagement. We need to control for the back door path to prevent a biased estimate of the effect of targeting on churn, which could even get the wrong sign. For example, targeting could mistakenly seem as if it increases churn rather than the real effect of decreasing it (also known as the Simpson Paradox). In general, this means that sufficient common causes of both the treatment and the outcome need to be included in the analysis, referred to as backdoor adjustment or controlling for confounding. Other biases could occur, including collider bias, selection bias, and mediators, see \cite{hernan2020causal, pearl2018book} for more details. 

\subsection{Types of Prescriptive Modeling}
Prescriptive Analysis holds within it a vast number of modeling frameworks. In this paper, we cannot review these methods fully and direct the reader to the citations. Nevertheless, we would like to give the reader a quick overview to help guide managers and practitioners. In general, there are three sources for prescriptive analysis: 
\begin{itemize}
    \item \textbf{Experimentation} - Randomly try actions to determine the cause and effect relationship. This could be a plain A/B test or more advance \textit{online} methods such as multi-arm bandits (MAB) and contextual bandits (CMAB), where the latter allows also personalization \cite{DBLP:journals/corr/abs-1904-10040, sawant2018contextual, ISSAMATTOS201968}. Online reinforcement learning (RL) is used when a sequence of related actions needs to be optimized.
    \item  \textbf{Simulation} - if a trustworthy simulator of the environment is available, such as in the case of video games \cite{DBLP:journals/corr/MnihKSGAWR13} or operational research \cite{BOUTE2022401}, optimization can be done using black-box optimization techniques \cite{reeves1997genetic}, simulated experimentation methods, online RL interacting with the simulation and a Monte-Carlo Tree Search (MCTS) approach \cite{6145622}.
    \item \textbf{Offline Data} - Observational studies\cite{pearl2018book} provide ways to leverage existing data to come up with better policies, mainly based on causal inference and recent major advances in offline RL~\cite{levine2020offline}. This is also useful when experimentation and simulation are not achievable.  Moreover, recommendation systems (RecSys) are often used when the size of the action space is very large \cite{wang2020causal}. One technique from the causal inference domain is Uplift Modelling, detailed in \ref{sec: uplift}.
\end{itemize}

In the next section, we will elaborate further on causal inference and observational studies as these are proven to be successful in practice for Marketing and Sales (see Section \ref{sec:benefit} quotes and references within).

\section{Causal Inference} 
\subsection{The Benefits of Causal Inference}
\label{sec:benefit}
Causal Inference is the set of techniques that allow inferring the actual causal effect of an action on the outcome \cite{yao2021survey, guo2020survey, pearl2018book}. The importance of causal inference was recently publicly recognized as it was at the center of the 2021 Economics Nobel prize \cite{hermann2022answering}. It is considered one of the top ten challenges of DS \cite{Wing2020Ten}. The causal framework plays an important role at leading enterprises, for example:

\begin{itemize}
    \item \textbf{Linkedin} - ``At LinkedIn, we proved the usefulness of observational causal studies to the business ... ''\cite{bojinov2020importance}.
    \item \textbf{Uber} - ``
    We have found it invaluable to bring causal inference methods to our work at Uber, as it enables us to solve challenging but critical data science questions that would otherwise be impossible to tackle...''\cite{harinen2019using}.
    \item \textbf{Netflix} - ``When A/B experiments are not possible, we rely on quasi-experiments and causal inference methods, especially to measure new marketing and advertising ideas.''\cite{NetflixR13}.
    \item \textbf{Amazon} - ``Focusing on causal effect leads to better return on investment (ROI)...''\cite{sawant2018contextual}.
\end{itemize}

\subsection{Necessary Assumptions for Causal Inference}
Applying causal inference on offline data requires three necessary assumptions~\cite{hernan2020causal}. These conditions must hold to conceptualize an observational study as a conditionally randomized experiment~\cite{hernan2020causal}:
\begin{enumerate}
    \item \textbf{SUTVA} - The values of treatment under comparison correspond to well-defined interventions, and the treatment of one unit does not interfere with other units. It also implies that outcomes of actions are consistent.
    \item \textbf{No Hidden Confounding} - The conditional probability of receiving every value of treatment depends only on the measured covariates, formally known as \textit{Ignorability}.
    \item \textbf {Positivity} - The probability of receiving every value of treatment was greater than zero for every decision.
\end{enumerate}

\textbf{SUTVA} means no hidden spillovers: the potential outcomes for a given unit are not influenced by actions on other units. The same action on the same unit is assumed to result in the same outcome. The \textbf{"No Hidden Confounder"} requirement implies that sufficient information should be available about the factors that affect both the treatment and the outcome are known. For example, if a targeting email is sent mainly to people who intend to buy, the effect of the email will be overestimated if it is not adjusted for that intention (see\cite{bojinov2020importance} for concrete numerical examples). The \textbf{positivity} requirement implies that every action probability is larger than zero for every decision. For example, if we never send emails to males, we can't determine the counterfactual predictions of emails for males.

\subsection{Uplift Modeling} \label{sec: uplift}
\begin{figure}[!htbp]
    \centering
    \includegraphics[width=0.35\textwidth]{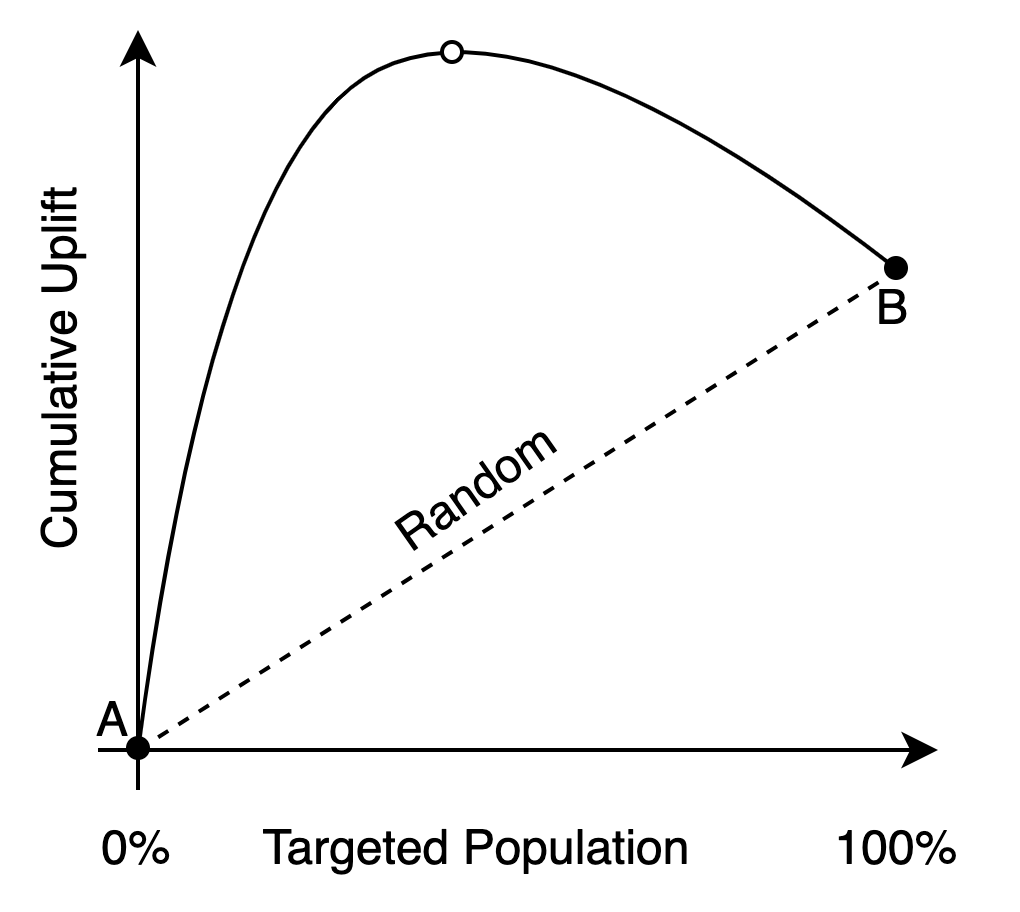}
    \Description{An uplift curve, with "Targeted Population" on the x-axis, and "Cumulative Uplift" on the y-axis. "Targeted Population" runs from 0\% to 100\%. A diagonal line, annotated as "Random", runs from zero (for both x and y, annotated as A) to 100\% targeted population and a positive cumulative uplift (annotated as B). An upside-down-U curve also runs from A to B, with the peak higher than B. }
    \caption{A schematic example of an uplift curve. The cumulative uplift as a function of the targeted population percentage. The optimal policy is to target only part of the population (empty circle) whereas an A/B test will test only the extremes 0\% (0) and 100\% (1) (filled circles)} 
    \label{fig:uplift}
\end{figure}
One specific useful approach to causal inference is uplift modeling \cite{zhang2021unified, olaya2020survey, devriendt2018literature}. Uplift modeling originates from a typical evaluation metric in marketing: the expected difference in outcome between the treated group and the control group (when the action is binary), also known as uplift \cite{surry2011quality}. In the simplest version, one determines the average treatment effect (ATE) of a whole population, but usually, uplift modeling estimates the individual treatment effect (ITE), also known as conditional average treatment effect (CATE). The ITE allows optimal target audience selection: target only people whose predicted effect is positive \cite{diemert2018large}. If constraints exist such as costs of targeting, one can use optimization techniques to find the overall optimal policy based on the ITE estimations~\cite{goldenberg2020free}. Determining the uplift performance for many different effect thresholds results in an uplift curve as shown in Figure~\ref{fig:uplift}. The two extreme points represent treating no one (A) vs treating everyone (B), two options in a common A/B test. As can be seen, the optimal policy is neither A nor B but somewhere in between. The ITE predictor can improve results compared to treating everyone (B) or no one (A) by utilizing the heterogeneity of the population. Notice that most uplift literature is focused on estimating ITE based on randomized control trials (RCT). To deal with observational data one needs to correct for biases \cite{gutierrez2017causal}. 

\section{The Prescriptive Modeling Canvas}
\begin{figure*}[!hbtp]
  \centering
  \includegraphics[width=0.85\paperwidth,angle=-90]{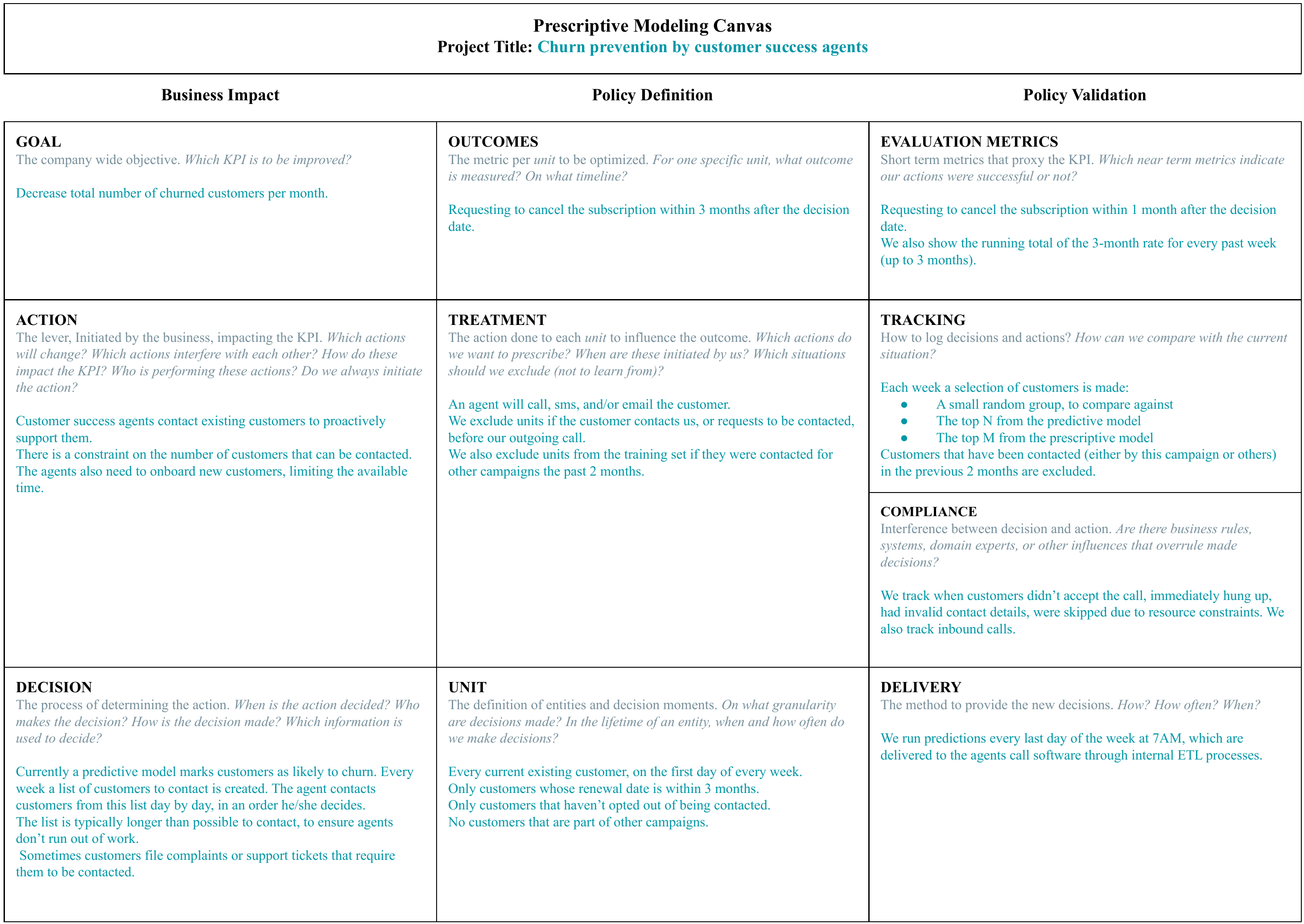}
  \caption{The Prescriptive Modeling Canvas. Example for churn prevention use-case is in Cyan.}
  \Description{A full page canvas, titled "Prescriptive Modelling Canvas", with subtitle "Project title: Sales Call optimization". It contains 3 columns: business impact, policy definition, and policy validation. The business impact column contains 3 cells: goal, action, and decision. The policy definition also contains 3 cells: outcomes, treatment, unit. The policy validation column contains 4 cells: evaluation metrics, tracing, compliance, and delivery.}
  \label{fig:our_canvas}
\end{figure*}

In the previous sections, we've elaborated on aspects and requirements of prescriptive modeling and the related techniques. It may seem daunting to design a successful prescriptive project. Therefore, we propose a canvas, as shown in Figure~\ref{fig:our_canvas}, to guide conversations between data scientists and business experts. It provides them with a structured checklist, that ensures important information is collected and easily verified. 

The canvas follows the recommendation of Microsoft's Team Data Science Process (TDSP)~\cite{marktab} regarding the first stage of ``business-understanding''. It also expands on the ``Drivetrain Approach'' explained in ``Designing Great Data Products''\cite{howard2012designing}, as it emphasizes the levers (actions) that truly modify business results. It also has some similarities with the popular checklist template named the ML canvas\cite{mlc}, by focusing on a clear problem definition from the data scientist's perspective. However, we combine the business levers with a prescriptive problem definition including actions and policies and add a clear evaluation plan. The canvas drives questions along the following three dimensions:
\begin{enumerate}
    \item \textbf{Business impact}: The scope of the project, with a definition of success.
    \item \textbf{Policy definition}: A specification of the problem in such a way that prescriptive methods can be applied.
    \item \textbf{Policy validation}: Applying and validating the resulting policy.
\end{enumerate}

\subsection{Business Impact}
The first column in the canvas ensures that the goal of the project, from a business perspective, is clear. It requires (1) a clear metric to optimize, (2) an action that we control, and (3) a decision process.

The \textbf{key performance indicator} (KPI) is the single, aggregated number, to be optimized by the newly created policy. This single metric, also known as the Overall Evaluation Criteria (OEC), forces trade-offs to be made between possible multiple metrics and aligns the organization behind a clear objective \cite{kohavi2020trustworthy}. 

The \textbf{action} is the lever that impacts the KPI. To determine a policy, we need to know which actions can be prescribed. While we put a lot of focus on the action, the policy can only perform a decision, which might not always lead to the decided action due to interfering processes. Investigating that interference allows us to adjust for biases and distinguish between good decisions and good actions. 

The policy will change, or replace part of, the current \textbf{decision} process. We first need to know how to integrate our optimal policy. We can also discover which information currently is used in the decision process. These probably make good features for the policy model. They can also indicate possible sources of bias, which will need to be adjusted for.

\subsection{Policy Definition}
The \textbf{Policy Definition} column maps the business perspective to a data perspective. The resulting formal policy definition allows us to apply prescriptive techniques. Here we construct explicit definitions for (1) units, (2) treatment, and (3) outcomes. 

Based on the analysis of the \textit{Decision} we can define the \textbf{unit}, which is a combination of an entity (like a customer, website visit, subscription) and a decision-moment, the point in time the action is decided.  It implies the granularity of the policy, how to define the rows for modeling, and explicitly states the eligibility: which units are included, and which are not.

The policy version of \textit{Action} is the \textbf{treatment}. Firstly it specifies which actions the policy can prescribe at which granularity. For example, to include calls and emails, but not SMS. But also whether to group them into one treatment "to contact", or to separate them into different actions. Secondly, it specifies how to map logged observational data to actual treatments. Based on the influences analyzed in the \textit{Action} section, one might want to exclude certain actions. For example, one can regard calls as treatments, but exclude calls when the customer scheduled such a call. Thus the treatment definition specifies for each unit, which action was observed, and whether the unit can be included.

While the \textit{KPI} was a single aggregate value, the \textbf{outcomes} defines the metric for each unit that needs to be optimized. If the KPI would be the churn rate, the outcome could be "churned within a month after we decided to call or not". Note that the outcome is defined as a measurement through time, starting at the decision-moment, for each unit.

\subsection{Policy Validation}
By specifying the policy validation approach, one firstly ensures that the policy actually can be validated. It will also identify technical work that needs to be done to measure the policy quality, which often can be started before the optimal policy has been created. 

First, we translate the \textit{outcome} to measurable \textbf{Evaluation Metrics}. In the churn case, one might use a 3-month horizon in the policy definition but evaluate with a short-term proxy "churned within 1 month" to reduce delays. The data scientist can verify the relationship between the proxy and the true outcome.

One needs to prepare \textbf{Tracking}, such that taken decisions and actions can be linked to the policy output.

In \textbf{Compliance} we will verify how performed actions correspond to taken decisions. Compliance will not be 100\%, as indicated by the previously specified interference. Therefore, the effect of each expected influence should be measured. This allows to verify the expectations, and one might be able to discover other not yet identified processes that also interfere.

The \textbf{Delivery} of decisions also needs to be prepared, such that the end-user can act on the prescribed actions. How and when will the policy be applied to new data. Is all the data available at that moment (instead of being logged with a delay)? How will the end-user be able to consume the decisions, and through which systems?

\section{Summary}
In this paper, we point out that the predictive approach is the common one in data science, even though it is not the optimal framework for optimizing business goals. We then present the prescriptive approach and emphasized its benefit in optimizing policies which in turn provides better business value. We also explained the challenges in the prescriptive approach, together with an overview of different approaches for prescriptive modeling. Causal inference and specifically uplift modeling were highlighted, as we believe these are underutilized, even though they represent three out of the top ten (3/10) essential most practical methods in DS \cite{Lo2020Top} (Sections 4.8, 4.1 and 4.2, respectively). We finalize the paper with a ``Prescriptive Canvas'' which aims at helping data science managers, product managers, and practitioners in framing ML projects from a prescriptive perspective, which we strongly believe can help the project's success probability. 

\bibliographystyle{ACM-Reference-Format}
\bibliography{bib}



\end{document}